\documentclass[conference]{IEEEtran}
\IEEEoverridecommandlockouts
\usepackage{cite}
\usepackage{amsmath,amssymb,amsfonts}
\usepackage{algorithm, algorithmic}
\usepackage{graphicx}
\usepackage{textcomp}
\usepackage{xcolor}
\usepackage{siunitx}
\usepackage{hyperref}
\usepackage{graphicx}
\usepackage{subfigure}
\usepackage{xfrac}
\usepackage{verbatim}
\def\BibTeX{{\rm B\kern-.05em{\sc i\kern-.025em b}\kern-.08em
    T\kern-.1667em\lower.7ex\hbox{E}\kern-.125emX}}

\begin{document}

\title{Double Deep Q Networks for Sensor Management in Space Situational Awareness}

\author{\IEEEauthorblockN{Benedict Oakes}
\IEEEauthorblockA{\textit{CDT in Distributed Algorithms} \\
\textit{University of Liverpool}\\
Liverpool, United Kingdom \\
sgboakes@liverpool.ac.uk}
\and
\IEEEauthorblockN{Dominic Richards}
\IEEEauthorblockA{\textit{Artificial Intelligence Group} \\
\textit{STFC Hartree Centre}\\
Daresbury, United Kingdom \\
dominic.richards@stfc.ac.uk}
\and
\IEEEauthorblockN{Jordi Barr}
\IEEEauthorblockA{\textit{Cyber and Information Systems Division} \\
\textit{Defence Science and Technology Laboratory}\\
United Kingdom \\
jmbarr@dstl.gov.uk}
\and
\IEEEauthorblockN{Jason Ralph}
\IEEEauthorblockA{\textit{Electrical Engineering and Electronics} \\
\textit{University of Liverpool}\\
Liverpool, United Kingdom \\
jfralph@liverpool.ac.uk}
}

\maketitle

\begin{abstract}
We present a novel Double Deep Q Network (DDQN) application to a sensor management problem in space situational awareness (SSA). Frequent launches of satellites into Earth orbit pose a significant sensor management challenge, whereby a limited number of sensors are required to detect and track an increasing number of objects. In this paper, we demonstrate the use of reinforcement learning to develop a sensor management policy for SSA. We simulate a controllable Earth-based telescope, which is trained to maximise the number of satellites tracked using an extended Kalman filter. The estimated state covariance matrices for satellites observed under the DDQN policy are greatly reduced compared to those generated by an alternate (random) policy. This work provides the basis for further advancements and motivates the use of reinforcement learning for SSA. 
\end{abstract}


\section{Introduction}

In an era of regular and frequent launches of satellites to low Earth orbit (LEO), the possibility of collisions between resident space objects continues to increase, and poses a significant threat to space based infrastructure. The Kessler Syndrome -- a cascade of collisions that could render any satellite use in LEO extremely difficult and costly -- becomes an increasing risk \cite{b1}. Between one third and one half of the capacity of LEO space has already been occupied \cite{b2}.

Space Situational Awareness (SSA) is the understanding of the complex orbital domain, involving man made objects, and natural phenomena \cite{b3}. Ground-based surveillance and tracking of man made objects in orbit can be achieved with a variety of instruments, including radars and optical telescopes. Measurements are required to be able to predict the trajectories of the objects, to assess the risk of potential collisions. However, measurements must be made repeatedly, as the orbit of any satellite is subject to change. These changes may be small perturbations, but the accumulation of small changes over time can be significant. In the LEO environment, there are several factors that could affect the orbit of satellites - most notably intentional manoeuvres, atmospheric drag, or solar radiation pressure could alter the orbit from a predicted trajectory. With limited sensor availability, efficient sensor management (SM) algorithms are necessary for long-term SSA. Given the large number of objects in LEO, the problem suffers from a combinatorial explosion as the number of possible actions increases \cite{b4}. The European Space Agency is investing in improving the long-term sustainability of the space domain \cite{b5}, and employing novel methods to improve SSA and help accomplish this goal. Objects orbiting the earth in LEO have short orbital periods, meaning they cannot be observed reliably from a single site; and these sites are often constrained to making measurements in clear weather and of restricted patches of sky. Therefore, using multiple sensors located around the globe is highly beneficial, but comes with a cost and is a considerable SM challenge.

Deep reinforcement learning (DRL) is one possible solution to this problem. DRL is the combination of standard reinforcement learning algorithms with neural networks to solve Markov decision processes (MDPs). DRL has been applied to various fields with large action spaces, and has produced impressive results \cite{b6}-\cite{b8}. 

\section{Filtering and State Estimation}
In this paper, we aim to estimate $\{X\}$, the set of state vectors describing satellites' positions and velocities, using measurements $\{Y\}$. The optimal state estimation algorithm for linear, Gaussian systems is the well known Kalman filter (KF). The Kalman filter is the best linear estimator for reducing the mean square error \cite{b9}. For slightly nonlinear systems, adaptations of the KF exist to attempt state estimation while overcoming some of these non-linearities. The extended Kalman filter (EKF) employs state transition and measurement functions, as opposed to simple matrices, to propagate the state estimates. However, the covariance is propagated linearly through the step, so the EKF is only suitable for systems with modest non-linearities. The unscented Kalman filter (UKF) develops this further, by generating sigma points around the target position, and propagating these through the non-linearity, and reforming the covariance \cite{b10}. In this paper we will only use an EKF for simplicity but the approach is readily extendable to UKF or other state estimation methods.

\section{Reinforcement Learning}
Reinforcement learning (RL) is a machine learning method in which an intelligent agent must make decisions to maximise its received reward, which is determined by the results of the actions it takes \cite{b11}, \cite{b12}. RL differs from other machine learning learning areas in that the model can be unknown, the agent need only know the actions and the reward, as well as some observation about the environment's transition into new time steps, based on the environment's evolution over time. Observations are usually related to some value in the environment that determines the amount of reward returned. This can be ideal for SM applications, particularly in SSA, where we do not need to model a potentially complex environment for the agent to interpret. This means RL can work in much higher dimensions than other dynamic programming approaches. 

Markov decision processes (MDPs) are the underlying formulations that RL algorithms are built upon. MDPs operate discretely, where at each time step, an action is made. The state will react to this action via a transition, and a reward is given. The transition is defined as $\mathcal{P}_{ss'} = \mathbb{P}[S_{t+1}=s'|S_t=s]$, where $\mathcal{P}_{ss'}$ is the state transition probability, $s$ is the Markov state at time $t$ and $s'$ is the successor state. The goal of an MDP is to find a policy, matching states to actions, to receive the maximum reward. 

Previous work into RL for SSA includes applications of DRL in \cite{b13} and \cite{b14}. They show proof-of-concept results for applying RL to the SSA problem, using Actor-Critic methods. The actor refers to the policy, that asks the estimated value function, or critic, about the next possible state values, which the critic improves during learning. More recently, extensions to the previous work were completed that added more complexity and required less intensive compute resources to run \cite{b15}, \cite{b16}. We present the first implementation of a DDQN to the sensor management problem for SSA, as opposed to the Actor-Critic methods cited above.

\subsection{Q-learning}

Q-learning is a simple value iteration update on a Markov decision process. Q-values, or quality-values, are state-action values, and refer to the expected reward gained by taking a certain action in a given state. Q-learning attempts to first find Q-values for a range of states and actions, and to then exploit the Q-values by selecting the action that returns the highest reward at any state, in a greedy policy. Q-learning is different from a Q-value iteration algorithm, as the transition probabilities and rewards are initially unknown. 

Q-learning is defined by: 
\small \begin{equation}
    Q^{new}(s_t,a_t) \leftarrow Q(s_t,a_t)+\alpha\cdot(r_t+\gamma\cdot\max Q(s_{t+1},a)-Q(s_t,a_t))
\end{equation} \normalsize
where Q is the expected reward and is a function of action $a$ and state $s$ at time $t$. $\alpha$ is the learning rate, and $0<\gamma<1$ is the discount factor. $\alpha$ is a tuning parameter that determines how quickly the algorithm learns new information. $\gamma$ is a parameter required for convergence of the algorithm, and determines how much weight is given to information in the future. If $\gamma$ is close to 1, the future is valued almost as much as the present. If $\gamma$ is close to 0, the immediate information is much more highly valued \cite{b17}. 

\subsection{Deep Q Network}
 A DQN is an implementation of Q-learning. The main issue with Q-learning is that it does not scale to larger problems with larger action and state spaces. Deep Q-learning was developed to overcome this. Deep Q-learning uses neural networks (NNs) and experience replay to use a random sample of prior actions instead of just the most recent action. Some well-known DQNs use convolutional NNs: hierarchical layers of tiled convolutional filters to mimic the effects of receptive fields \cite{b18}. Receptive fields are defined as the association of input fields to output fields. Experience replay is the use of batches of sampled transitions for better data efficiency and stability. DQNs are the term given to implementations of Q-learning algorithms applied to such NNs. 

\subsection{Double Deep Q Network}
It has been shown that DQNs commonly overestimate action values in certain situations, and produce over-confident Q-values \cite{b19}. To solve this problem, Double Deep Q Networks were developed. In DQNs, the max operator is used to select and evaluate actions, which leads to overly confident value estimates. By using two sets of weights $\theta$ and $\theta'$, and using one to determine the policy and the other to evaluate it, this problem is effectively overcome. 

\section{Problem Simulation}
\begin{figure}[htbp]
    \centering
    \centerline{\includegraphics[width=.45\textwidth]{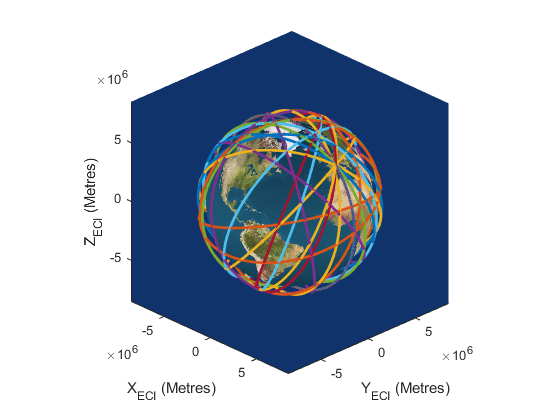}}
    \caption{LEO satellite paths plotted around the earth in the ECI reference frame}
    \label{fig:globe_orbits} 
\end{figure}

In this scenario, we create a satellite simulation using a Python package Pysatellite: a Github repository being developed by the author \cite{b24}. This package implements orbit generation, reference frame transformations, and target tracking -- in this scenario through the use of an EKF. We generate 25 LEO satellites using a Keplerian model, visualised in Fig.~\ref{fig:globe_orbits}. Higher order terms such as solar radiation pressure and atmospheric drag will be included in further iterations. In this paper, it is assumed that all satellites are following circular orbits at a radius from the centre of the Earth of $R = \num{7e6}$ metres. For this implementation, we find that using an EKF is adequate to handle the non-linearities of the system, but in future work, a UKF or particle filter may be more suitable.

\begin{figure}[htbp]
    \centering
    \centerline{\includegraphics[width=.45\textwidth]{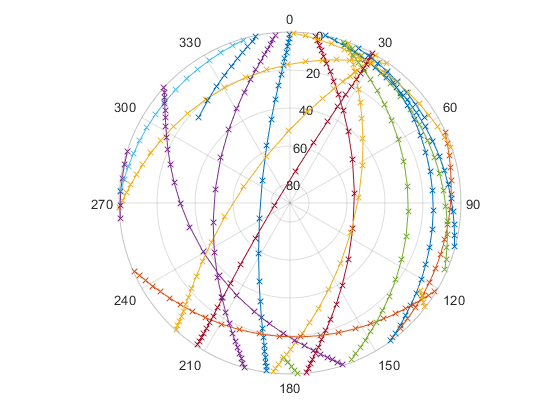}}
    \caption{Satellite paths crossing an Earth based telescope's Field of View}
    \label{fig:teles_fov} 
\end{figure}

Detections are generated by simulating a telescope on the surface of the Earth which measures azimuth, elevation, and range with additive Gaussian noise. Fig.~\ref{fig:teles_fov} shows the satellite paths from the telescope's point of view. The measurements are transformed into the Earth-centred inertial (ECI) reference frame, a Cartesian frame with the origin at the centre of the Earth, through which the Earth rotates. The EKF operates in an ECI reference frame, with a state vector $X = (x, y, z, x_v, y_v, z_v)$, which encodes the Cartesian position and velocity of the satellite. Measurements are transformed from azimuth, elevation, and range to ECI via the following method:

\begin{equation}
    Y_{AER} = \begin{bmatrix}
        \phi \\ 
        \theta \\ 
        R 
    \end{bmatrix}
\end{equation}
\begin{equation}
    Y_{NED} = \begin{bmatrix}
    R \cdot \cos(\theta) \cdot \cos(\phi)\\
    R \cdot \cos(\theta) \cdot \sin(\phi)\\
    -R \cdot \sin(\theta)
    \end{bmatrix}
\end{equation}

\begin{equation}
\begin{split}
    Y_{ECEF} &= \begin{bmatrix}
    -\sin(\phi_0)\cos(\lambda_0) & -\sin(\lambda_0) & -\cos(\phi_0)\cos(\lambda_0) \\
    -\sin(\phi_0)\sin(\lambda_0) & \cos(\lambda_0) & -\cos(\phi_0)\sin(\lambda_0) \\
    \cos(\phi_0) & 0 & -\sin(\phi_0)
    \end{bmatrix}\\ 
    &\cdot Y_{NED}
\end{split}
\end{equation}

\begin{equation}
\begin{split}
    Y_{ECI} &= \begin{bmatrix}
    \cos(\omega t) & -\sin(\omega t) & 0 \\
    \sin(\omega t) & \cos(\omega  t) & 0 \\
    0 & 0 & 1
    \end{bmatrix}\\ 
    &\cdot Y_{ECEF}
\end{split}
\end{equation}
\noindent where $\phi$, $\theta$, and $R$ are the azimuth, elevation and range coordinates of the satellite respectively, $Y_{NED}$ refers to the commonly used North, East, Down reference frame, $\phi_0$ and $\lambda_0$ are the latitude and longitude positions of the sensor, respectively, $\omega$ is the Earth rotation rate, and $t$ is the sidereal time. In our simulated problem, we define our own ECI and ECEF reference frames related to the time elapsed between frames. For dealing with real data, ECI and ECEF reference frames that relate to common time bases must be used. The method explained here does not account for higher-order specificities associated with real-world reference frames, but is suitable for a simple geoid simulation. 

A measurement noise matrix is generated in the AER frame for each measurement. We simulate ideal diffraction limited measurements. 

\begin{equation}
    P_{AER} = \begin{bmatrix}
    \sigma_\theta^2 & 0 & 0 \\
    0 & \sigma_\theta^2 & 0 \\
    0 & 0 & \sigma_r^2
    \end{bmatrix}
\end{equation}
\noindent where $\sigma_\theta$ and $\sigma_r$ are the standard deviations expected for ideal angle and range measurements. These are converted to the ECI frame by calculating the Jacobian matrix of the measurement through the above transformation from AER to ECI and applying it to the measurement noise matrix:

\begin{equation}
    P_{ECI} = J \cdot P_{AER} \cdot J^T
\end{equation}
\noindent where $J$ is the calculated Jacobian matrix. \\

For the learning environment, we use Tensorflow Agents, an accessible and approachable RL framework in Python. Agents allows users to create their own environments, and apply them to a range of different RL algorithms, including DQN, Deep Deterministic Policy Gradient (DDPG), and others; for our simulation we used a DDQN. The DDQN uses a replay buffer and stochastic gradient descent to calculate the loss. 

We state the following definitions for clarity: iterations refer to the number of training episodes that have occurred. Episodes refer to one full run of an environment, made up of $t$ time-steps. At each iteration, the step transitions are added to a circular buffer that stores the last $n$ number of iterations. During learning, a small sample of the buffer is used to calculate the loss, instead of just the last transition. This provides two benefits: as each transition is sampled many times, a higher data efficiency is achieved, and using uncorrelated transitions leads to a better stability of data. 

We model a telescope control scenario, with a sensor state $T = (\phi, \theta)$, where $\phi$ is the telescope's azimuth pointing, and $\theta$ is the telescope's elevation pointing. At each time step of the environment, the agent can choose from 5 possible actions: move up, down, left, right, or do nothing. As the environment is discretised, we assume that when an action is taken, the telescope state in the next time-step will be equal to the maximum distance travelled in that direction, based on a telescope slew rate of $2^o /s$. Up and down refer to the telescope's elevation pointing, and left and right refer to the telescope's azimuth pointing. If actions are taken that would be unfeasible, such as the telescope pointing below the horizon, no action is taken. As elevation can cross the zenith at $\sfrac{\pi}{2}^c$, and azimuth is bounded $0 < \theta < 2\pi^c$, actions that would take the telescope direction out of this range are wrapped. For each satellite, we find the difference between the centre of the telescope Field of View (FoV) and the satellite's azimuth and elevation position. 
\begin{align}
    d_{\phi} &= |s_{\phi} - T_{\phi}| \\
    d_{\theta} &= |s_{\theta} - T_{\theta}|
\end{align}
\noindent where $d_{\phi}$ and $d_{\theta}$ are the differences in azimuth and elevation, respectively, $s_{\phi}$ and $s_{\theta}$ are the satellite's azimuth and elevation positions, respectively, and $T_{\phi}$ and $T_{\theta}$ are the telescope's centre azimuth and elevation pointing, respectively. If both azimuth and elevation are within the FoV of the telescope, then a reward is given. The reward is cumulative in each time-step, so if multiple satellites are detected, the reward will increase accordingly. Algorithm 1 shows the basic operation of the RL environment.

\begin{algorithm}
\caption{RL environment}
\begin{algorithmic}
\STATE $t$ = time-step
\STATE $a$ = action
\STATE $r$ = reward
\STATE $s$ = satellite, $s_p$ = satellite position
\STATE $T(\phi, \theta)$ = sensor state with field of view FoV
\STATE $o$ = observation
\WHILE{episode not ended}
    \STATE $t$ += 1
    \STATE Get $a$ for current $t$
    \IF {$a <$ limit} 
      \STATE apply $a$
    \ELSE
        \STATE wrap and apply $a$
    \ENDIF
    \FOR {each $s$}
        \STATE $o = T(\phi, \theta) - s_p$
        \IF {$o$ within FoV}
            \STATE measurement = True
            \STATE $r$ += 1
        \ELSE
            \STATE measurement = False
            \STATE $r$ += 0
        \ENDIF
    \ENDFOR
    \RETURN $r$, $o$
    \IF {last $t$}
        \STATE end episode
    \ENDIF
\ENDWHILE
\end{algorithmic}
\end{algorithm}

Algorithm 2 shows the training loop to improve the agent's policy, and select the best action at each time-step. The agent will train for a certain number of iterations, and the performance is periodically evaluated using a sample of episodes which are executed with the current policy. We then run another episode to generate measurements using the current policy, which are used in the EKF. 

\begin{algorithm}
\caption{DDQN Training}
\begin{algorithmic}
\STATE $t$ = time-steps
\STATE $r$ = reward
\STATE $Y$ = measurements
\FOR {each iteration}
    \STATE collect multiple $t$, save to replay buffer
    \STATE sample buffer, update network
    \IF {iteration = evaluation interval}
        \STATE calculate $\overline{r}$ of 10 episodes with current network
        \STATE generate set of $Y$ from current network
        \STATE use $Y$ in EKF
    \ENDIF
\ENDFOR
\end{algorithmic}
\end{algorithm}

\section{Results} 
\begin{figure}[htbp]
    \centering
    \centerline{\includegraphics[width=.45\textwidth]{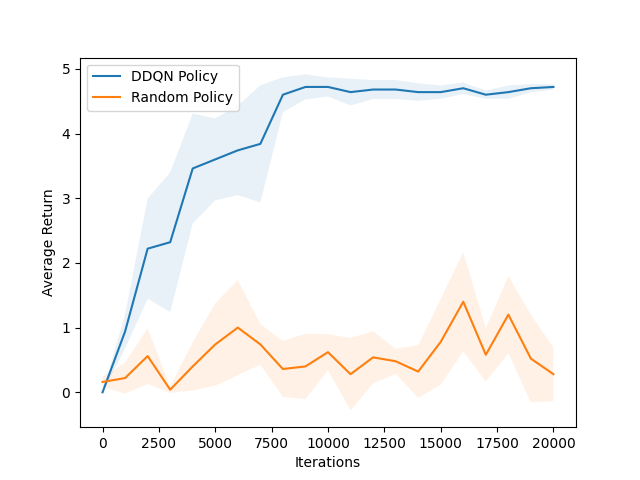}}
    \caption{Average reward from 10 episodes returned by an environment trained using a DDQN. The trained policy in blue shows vast improvements against the random policy. Shaded regions show the standard deviation of 5 training runs}
    \label{fig:TCRwA} 
\end{figure}

After training the above environment on a DDQN for 20,000 iterations, and sampling the average reward at every 1,000 iterations, Fig.~\ref{fig:TCRwA} shows that the DDQN agent clearly outperforms the same environment run with a random policy, where at a given time-step, an action is picked at random, instead of choosing the action that will maximise the cumulative reward. We choose to compare against a random policy to show a base-line for learning, and to show clearly the improvement of the DDQN over increasing iterations. Future work will compare the DDQN to other RL implementations. Each iteration trains the agent with 10 episodes of the environment, with each episode consisting of 20 time-steps. We use a seeded simulation to create the same satellite orbits, and run the environment for 5 sets of iterations to generate average returns. 

\begin{figure}[htbp]
    \centering
    \centerline{\includegraphics[width=.42\textwidth]{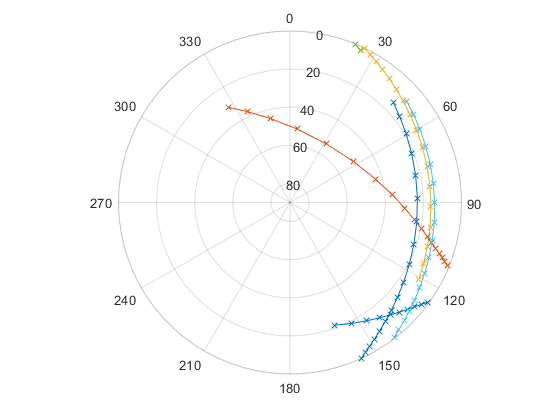}}
    \caption{Satellite paths crossing an Earth based telescope's FoV over a limited number of time steps - most satellites will not appear in telescope FoV}
    \label{fig:teles_fov_short} 
\end{figure}

It is clear that after $\sim$8,000 iterations, the DDQN begins to converge on an optimal policy, that far exceeds the random policy shown, which has no obvious improvement over time, as expected. After this, the DDQN remains at or near the optimal policy, with no loss of return from catastrophic forgetting, a common problem in RL algorithms \cite{b20}. Catastrophic forgetting occurs as an agent explores its environment, it may learn things that break its previously learnt information, causing the agent to forget the past information and return poor rewards. The shaded region shows the standard deviation of 5 runs of the same environment in the DDQN. The standard deviation decreases once the algorithm reaches the plateau, showing its increased confidence in this region of training. The maximum possible return of the DDQN is less than the number of satellites simulated because only some of the satellites will cross the telescope field of view during the length of the environment simulation, as exemplified in Fig.~\ref{fig:teles_fov_short}. Increasing the number of steps in each episode would lead to a higher maximum possible return, but training would take far longer due to the increased action space over the larger number of steps, and thus would require more iterations to converge. 

\begin{figure}[ht]
\centering
\begin{minipage}[b]{0.47\linewidth}
\includegraphics[width=\textwidth]{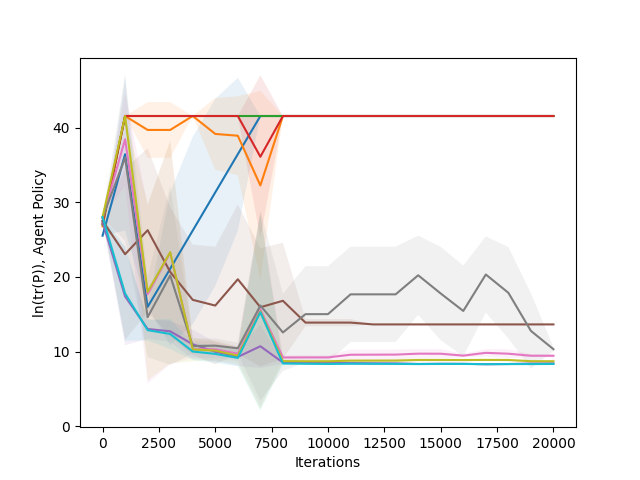}
\caption{Log-trace of final covariance matrices for tracked satellites at the end of an episode. Each line represents a different satellite visible in the telescope FoV. Shaded regions show the standard deviation of 5 training runs}
\label{fig:trace_cov}
\end{minipage}
\quad
\begin{minipage}[b]{0.47\linewidth}
\includegraphics[width=\textwidth]{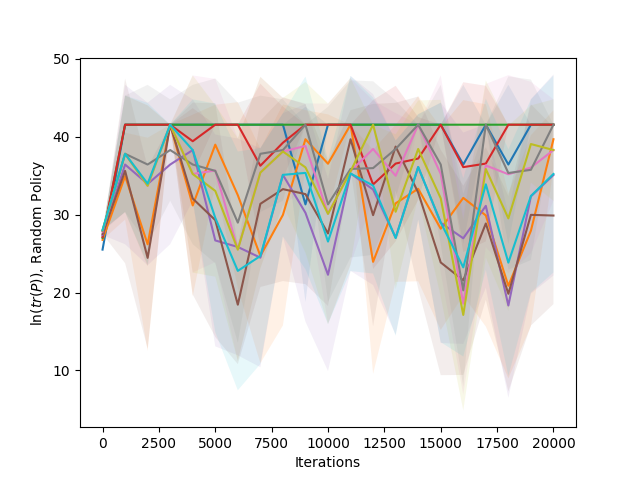}
\caption{Log-trace of final covariance matrices for tracked satellites at the end of an episode, with a random policy. Each line represents a different satellite visible in the telescope FoV. Shaded regions show the standard deviation of 5 training runs}
\label{fig:trace_cov_rand} 
\end{minipage}
\end{figure}

In Fig.~\ref{fig:trace_cov}, we show the log of the trace of the covariance matrices associated with each satellite after they have been tracked for the length of the episode. The value for each point in the graph is the trace of the covariance matrix after applying an EKF to the satellite for $n$ steps, where $n$ is the number of steps used in the RL environment, based on the measurements generated in the DDQN.

 If a satellite is captured within the telescope FoV, a measurement is generated; conversely if the satellite is not observed by the telescope, no measurement can be made. As the policy improves and the number of satellites seen in the FoV of the telescope increases, more measurements are generated as iterations increase. By having more measurements for each satellite, the EKF is able to reduce the uncertainty of the target position and velocity, which can be seen in the lines at the bottom of the graph. Where limited or no measurements are made, the EKF can only predict the satellite position and velocity, giving increasing uncertainty -- seen at the top of the graph. We see that over half of the visible satellites are measured more consistently as the DDQN trains, leading to reductions in the final uncertainty of the satellite. We compare this with Fig.~\ref{fig:trace_cov_rand}, which is the result of tracking on measurements generated from a random policy. Here we see no overall improvement on the tracking performance.

\begin{figure*}[ht]
\centering
\begin{minipage}[b]{0.45\linewidth}
\includegraphics[width=\textwidth]{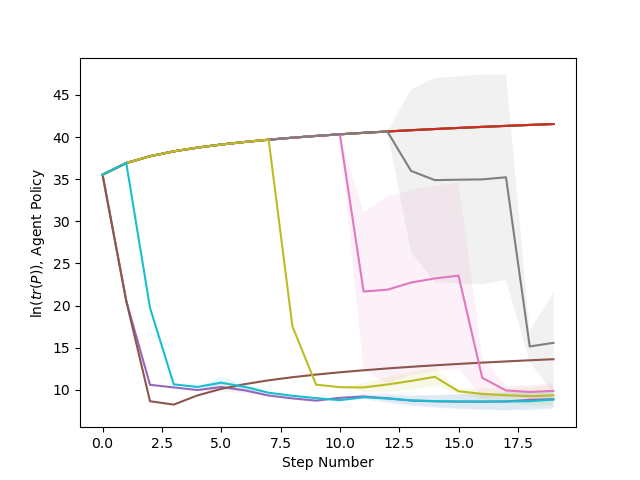}
\caption{Satellite covariances during the final trained episode. Each line represents 1 satellite. Shaded regions show the standard deviation of 5 training runs}
\label{fig:fin_cov}
\end{minipage}
\hfill
\begin{minipage}[b]{0.45\linewidth}
\includegraphics[width=\textwidth]{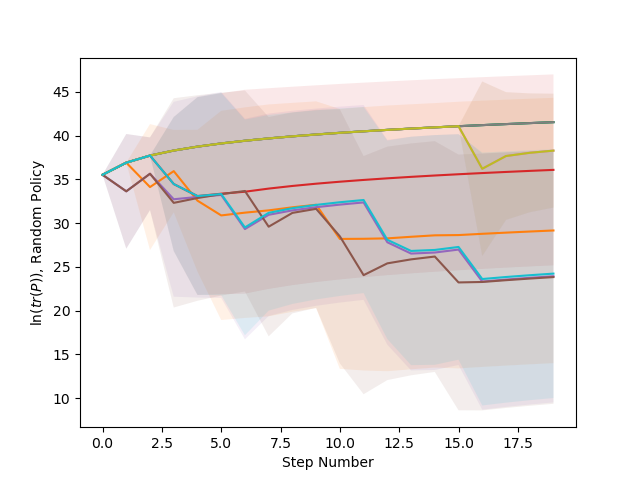}
\caption{Satellite covariances during the final random episode. Each line represents 1 satellite. Shaded regions show the standard deviation of 5 training runs}
\label{fig:fin_cov_rnd}
\end{minipage}
\end{figure*}

In Fig.~\ref{fig:fin_cov} and Fig.~\ref{fig:fin_cov_rnd}, we show the outcome of tracking in the final iteration, when the agent has attained the optimal reward. In the trained run, we see that a majority of the satellites are detected by the telescope, meaning the measurements made are able to reduce the uncertainty. Some satellites are never seen in the FoV, which is shown by the top line in the graph, where the EKF becomes increasingly uncertain about its state. In the random action run, we can see that no satellite has consistent measurements, meaning that the uncertainties are not lowered as well. This shows a SM algorithm that is better than random pointing, proved by a covariance-based metric.

\section{Future Work}
In future work, we hope to expand on the work completed here, with the inclusion of target tracking performance metrics. Two such metrics that would likely prove fruitful in this scenario are the Posterior Cramér-Rao Bound (PCRB) \cite{b21} and the Generalised Optimal Sub-Pattern Assignment (GOSPA) \cite{b22}. The PCRB would be useful in situations where the geometry affects the resulting information of the targets, such as in cases where the covariance of a target is very thin but long. The GOSPA metric will be more suitable in scenarios where there is clutter, false detections, and missed targets \cite{b23}, all of which are likely in the SSA domain. Further improvements will be made to increase the complexity of the satellite dynamics and how they are tracked, including more robust reference frame transformations, for example. Including effects like solar radiation pressure and atmospheric drag will increase the realism of the scenario, and will require more advanced tracking algorithms, such as a UKF. Other advancements include using angle-only measurement models, and continuous control agents to more accurately reflect the use of real telescopes. 

\section{Conclusions}
In this paper, we simulate a controllable Earth-based telescope viewing satellites in low Earth orbit in a reinforcement learning environment. We present a novel application of a Double Deep Q Network to space situational awareness. We maximise the number of satellites observed during a time period, and increase the number of successful measurements made. We use the generated measurements in an extended Kalman filter, in which we see a significant reduction in position and velocity uncertainty for observed satellites as a result of increasing observations, as opposed to observations made from a random policy. This forms the basis of a framework for future research into applying deep reinforcement learning to space situational awareness.

\end{document}